\documentclass{article}
\usepackage{spconf,amsmath,graphicx}
\usepackage{enumitem}
\usepackage{xcolor}
\setlist{nosep, leftmargin=14pt}

\begin{document}

\title{An Attention Based Pipeline for Identifying Pre-Cancer Lesions in Head and Neck Clinical Images\\
}
\name{Abdullah Alsalemi$^{a \mathsection}$ \qquad Anza Shakeel$^{b \mathsection}$ \qquad Mollie Clark$^{b}$ \qquad Syed Ali Khurram$^{b \star}$ \qquad Shan E Ahmed Raza$^{a \star}$}

\address{$^{a}$ Tissue Image Analytics Centre, Department of Computer Science, University of Warwick \\
$^{b}$ Unit of Oral and Maxillofacial Pathology, School of Clinical Dentistry, University of Sheffield \\
$^{\mathsection}$Joint first authors \hspace{1em} $^{\star}$Joint senior authors} 

\maketitle

\begin{abstract}
Early detection of cancer can help improve patient prognosis by early intervention. Head and neck cancer is diagnosed in specialist centres after a surgical biopsy, however, there is a potential for these to be missed leading to delayed diagnosis. To overcome these challenges, we present an attention based pipeline that identifies suspected lesions, segments, and classifies them as non-dysplastic, dysplastic and cancerous lesions. We propose (a) a vision transformer based Mask R-CNN network for lesion detection and segmentation of clinical images, and (b) Multiple Instance Learning (MIL) based scheme for classification. Current results show that the segmentation model produces segmentation masks and bounding boxes with up to 82\% overlap accuracy score on unseen external test data and surpassing reviewed segmentation benchmarks. Next, a classification F1-score of 85\% on the internal cohort test set. An app has been developed to perform lesion segmentation taken via a smart device. Future work involves employing endoscopic video data for precise early detection and prognosis.
\end{abstract}

\begin{keywords}
Cancer early detection, deep learning, segmentation, classification, head and neck cancer, dysplasia.
\end{keywords}

\section{Introduction}
\label{into}

Early detection of cancer can help boost patient survival as well as reduce treatment cost and duration \cite{nevanpaa_malignant_2022}. One of the commonest cancers is head and neck cancer, ranked as $8^{th}$ in the UK \cite{noauthor_head_2019} and $7^{th}$ globally \cite{gormley_reviewing_2022}, with rising frequency and poor prognosis. In the UK alone, about 12,000 cases are recorded yearly, with a striking fact that between 46\% and 88\% are preventable, partly through early detection of ``potentially malignant'' conditions \cite{noauthor_head_2019}.

This is where AI can play a pivotal role, i.e., in the early detection of head and neck cancer, e.g., Oral Cavity Squamous Cell Carcinoma (OSCC), by identifying common pre-malignant conditions such as Oral Epithelial Dysplasia (OED) in clinical images \cite{mahmood_prediction_2022}. In a general practitioner appointment, an OED lesion can be identified via its visual features including colour, texture, and shape. Such screening can be followed up by a specialist appointment to verify symptoms and malignancies with a biopsy. In this context, Deep Learning (DL) can speed up the triage and referral in either an invasive or non-invasive manner: (a) by processing clinical images taken by a GP or a dentist during a regular check-up \cite{camalan_convolutional_2021}; or (b) by analysing Whole Slide Images (WSIs) taken via biopsies of the lesion tissue, which is considered the gold standard, providing ever increasing reliability and breadth of available information \cite{shephard_fully_2023}; The former approach overcomes the limitations of an invasive biopsy by promising a quick pipeline that can identify and classify the lesion at an early stage, reducing waiting time and associated costs for collecting invasive biopsies.

Hitherto, a sparse amount of studies have been dedicated to OED detection, segmentation, and classification using clinical images compared to the computational pathology approach \cite{mahmood_artificial_2021}. In terms of segmentation, Anantharaman et al. \cite{anantharaman_utilizing_2018} utilise Mask R-CNN to segment oral cancer lesions from an online-sourced dataset of forty images annotated by the authors, with an average Dice score of 68.3\%.
Also, Tanriver et al. compiled 684 images to create a detection, segmentation, and classification system of head and neck cancer. A Dice score of 86.0\% using a U-Net model is obtained and a classification F1-score of 86.0\% is achieved using an EffcientNet-B4 network.

One of the most prominent contributions in lesion classification is Fu et al. \cite{fu_deep_2020}, where a dataset of 44,409 clinical images are collected, pre-processed, augmented, and employed to fine-tune a DenseNet network. Binary classification results of OSCC results include an AUC of 98.3\%. It is noteworthy to mention that the a mobile app based implementation is presented as part of the evaluation study. 

However, many of the studies only look at cancer detection, differentiating it from non-cancer, which is a simpler and easier task and unfortunately do not take into account OED or potentially malignant lesions.

Therefore, in this paper, we present an OED detection, segmentation, and classification system that involves clinical images taken from multiple centres. Using a novel, annotated OED clinical image dataset from a partner university hospital, including more than 230 cases from an internal cohort and about 400 images from two external datasets, the main technical contributions are as follows:
\begin{enumerate}
    \item This is the first work to introduce attention based architectures into an OED detection, segmentation, and classification pipeline in clinical images:
\end{enumerate}
\begin{itemize}
    \item[(a)] an improved architecture based upon Mask R-CNN with vision transformers to detect and segment OED and cancerous lesions that is robust to varying image resolutions and image acquisition environments; and
    \item[(b)] lesion classification using batch based Multiple Instance Learning (MIL) with VGG-16 to grade lesions as non-dysplastic, dysplastic, and cancerous.
\end{itemize}
\begin{enumerate}
    \item[2.] Creating an app that showcases the segmentation model in action with the ability for a practitioner to upload images for OED lesion segmentation.
\end{enumerate}

The remainder of this paper is organised as follows. Section \ref{sec:methods} presents an overview of the detection, segmentation, and classification system as well as the employed dataset. Section \ref{sec:results} reports results and discusses limitations of the model. Section \ref{sec:conclusions} concludes the paper with a blueprint for future work.

\section{Methods}
\label{sec:methods}

\begin{figure}[!t]
\centering
\noindent\includegraphics[width=1.05\linewidth]{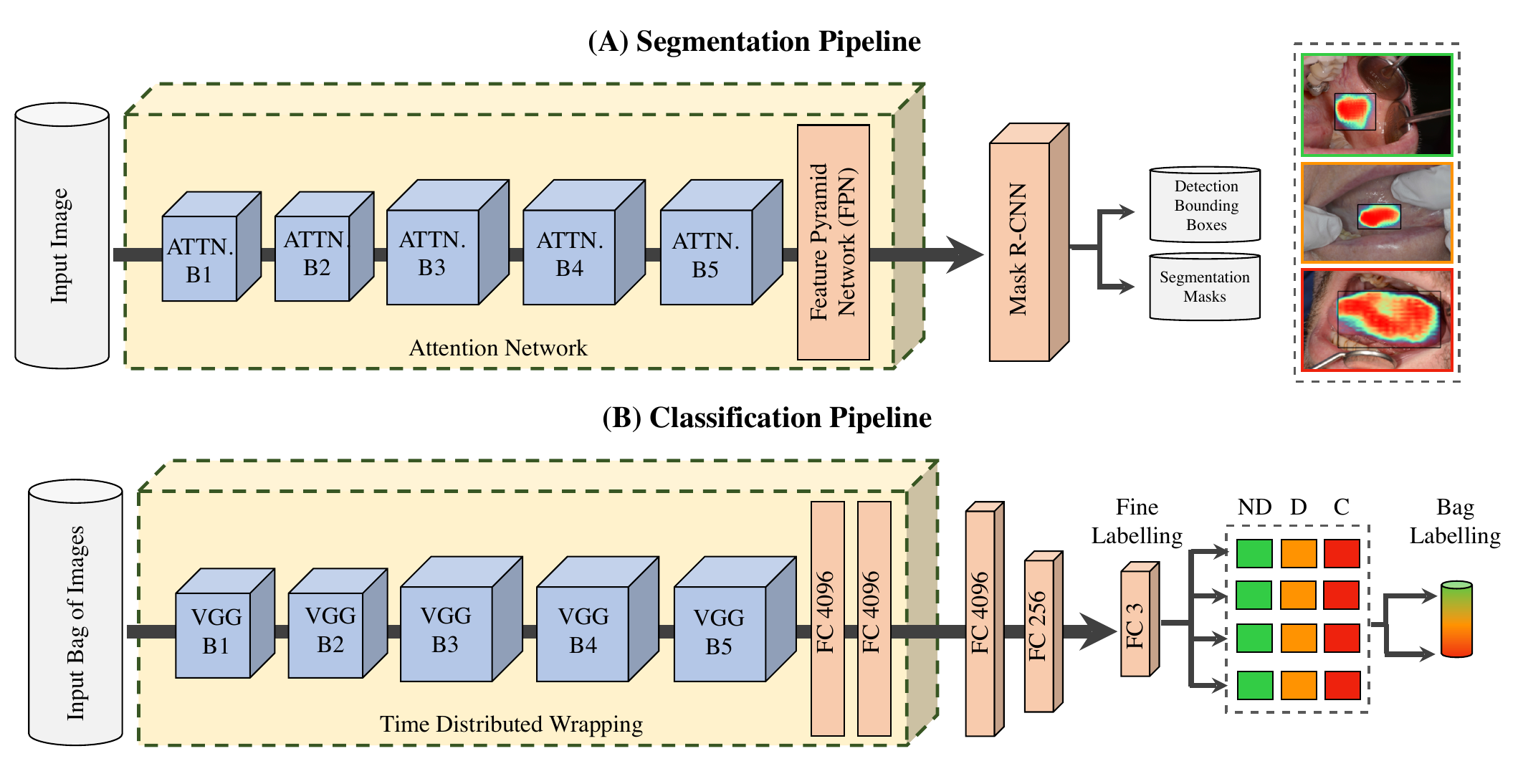}
\caption{Block diagram of the proposed (A) vision transformer based segmentation and (B) VGG-16 based Multiple Instance Learning (MIL) classification pipeline.}
\label{fig:block-diagram}
\end{figure}

In this section, we describe the proposed OED early detection system in three parts: (a) clinical image dataset, (b) detection and segmentation, and (c) classification. An overview of the proposed system is shown in Fig. \ref{fig:block-diagram}.

\subsection{Clinical Image Dataset}
\label{ssec:data}

\begin{figure}[!t]
\centering
\noindent\includegraphics[width=0.85\linewidth]{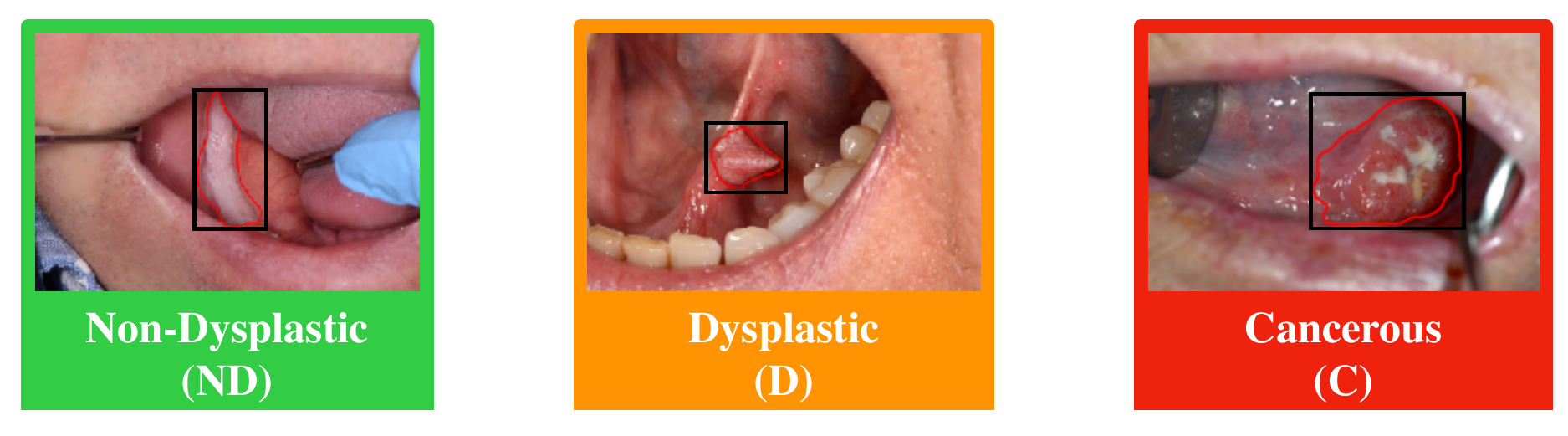}
\caption{\textcolor{black}{Sample of clinical oral photographs from the Sheffield cohort representing non-dysplastic (green), dysplastic (orange) and cancerous (red) lesions. Expert annotations of affected lesions are shown in red on each image.}}
\label{fig:data}
\end{figure}

In collaboration with University of Sheffield, clinical image data was collected from a cohort of 280 cases at the School of Clinical Dentistry. The clinical image based OED dataset is comprised of the followed classes: (a) cancerous (35 cases), (b) dysplastic (160 cases), and (c) non-dysplastic (85 cases). \textcolor{black}{The data has been obtained as part of a research project conducted at the said institution with non-OED, OED, and OSCC patients, where all collected images are biopsy proven to obtain ground truth diagnosis.}
All cases are anonymised, with an average image resolution of $5,247$ $\times$ $3,567$ px. 
Fig. \ref{fig:data} shows a sample of the cohort data.

For further model optimisation and testing, two external datasets are employed. Firstly, Barot et al.'s oral cancer dataset includes (a) cancerous ($87$ cases) and non-cancerous (44 cases), which is used for training. Secondly, Chandrashekar et al.'s oral cancer dataset includes (a) 158 cases cancerous, and (b) non-dysplastic ($142$ cases) used for testing. Image size varies between $93$$\times$$96$ px and $9,248$$\times$$6,936$ px, with an average of $756$$\times$$731$ px. It is noteworthy to mention the diversity of image resolutions of the collected data, which helps train the model for further robustness against different input image sizes. The dataset is used according to Creative Commons CC BY 4.0 license. Following image loading, a pre-processing stage is carried out to prepare the data for further analysis, which involves different data augmentation operations. \textcolor{black}{Such operations include horizontal flip, random rotation, Gaussian blur, random brightness contrast, and random hue saturation shift. Also, if the image is has a size more than 512$\times$512 px it is downsampled by a factor of three in order to speed up training. For the classification task, the internal cohort is split into 70\%, 20\%, and 10\% for training, validation, and testing respectively.}

\subsection{Lesion Detection and Segmentation}
\label{ssec:segmentation}

In order to segment OED lesions from a given clinical image, the first step is to identify the correct Region of Interest (ROI). Following this, segmentation is performed, where a mask is drawn around the border of the lesion, producing a bounding box and a segmentation mask. 

In this work, an enhanced attention based Mask R-CNN is employed to perform both the detection and segmentation tasks. Developed by Meta's Fundamental AI Research (FAIR) in 2017, Mask R-CNN is an advancement over Faster R-CNN, and can be used to perform instance segmentation using region proposals. The model extracts features using a Convolutional Neural Network (CNN) \cite{anantharaman_utilizing_2018} that are turned into bounding box region proposals through the Regional Proposal Network. Then an ROI Align network performs the segmentation task, producing a proposed segmentation for each extracted bounding box using a Fully Convolutional Network (FCN). The loss function of Mask R-CNN is defined in (\ref{equ:mask-rcnn-loss}).

\begin{equation}\label{equ:mask-rcnn-loss}
\mathcal{L}_\text{maskrcnn} = \mathcal{L}_\text{cls} + \mathcal{L}_\text{box} + \mathcal{L}_\text{mask}
\end{equation}

\begin{equation}\label{equ:box-loss}
\mathcal{L}_\text{box} = \frac{\lambda}{N_\text{box}} \sum_i p^*_i \cdot L_1^\text{smooth}(t_i - t^*_i)
\end{equation}

\begin{equation}\label{equ:mask-loss}
\mathcal{L}_\text{mask} = 1-(2y\widehat p+1)/(y+ \widehat p+1)
\end{equation}

\begin{equation}\label{equ:l1-loss}
L_1^\text{smooth}(x) = \begin{cases}
    0.5 x^2             & \text{if } \vert x \vert < 1\\
    \vert x \vert - 0.5 & \text{otherwise}
\end{cases}
\end{equation}

where \(\mathcal{L}_\text{cls}\) is classification loss defined as the cross entropy loss \cite{he_mask_2018}, and the regression bounding box loss is defined as (\ref{equ:box-loss}). The the revised mask loss \(\mathcal{L}_\text{mask}\) is defined in (\ref{equ:mask-loss}) as the Dice loss function \cite{jadon_survey_2020}, replacing the default binary cross entropy loss, where \(p\) is the predicted probability at a given pixel where \(p_i\) is the predicted probability of anchor \(i\) being a ROI; \(p^*_i\) is the ground truth label for \(i\) being a ROI; \(t_i\) is the predicted four parameterised coordinates; \(t^*_i\) is the ground truth coordinates; \(N_\text{box}\) is a normalisation term; \(\lambda\) is a balancing parameter; and \(L_1^\text{smooth}\) is the smooth L1 loss defined as (\ref{equ:l1-loss}). \textcolor{black}{We have employed the Dice loss due to its robustness in handling class imbalance and its sensitivity in detecting small ROIs, which are evident in the dataset.}

Moreover, an updated version of Mask R-CNN introduced in \cite{li_benchmarking_2021} involves the use of vision transformers. The networks employs non-overlapping windowed attention in order to optimise for computational time and memory (as shown in Fig. \ref{fig:block-diagram}A). Thus, the upgraded transformer based Mask R-CNN can achieve up to 4\% improvement in terms of average precision. The novelty of the proposed approach is the use of attention mechanism plus enhancing the architecture by using the Dice loss function to improve ROI detection and segmentation performance in order to produce accurate and reliable segmentation on unseen data. \textcolor{black}{In terms of parameters, a learning rate of 0.0001 and a weight decay 0f 0.0005 are employed in fine-tuning the network on a ResNet-50 Feature Pyramid Network (FPN) backbone.}

\subsection{Lesion Classification}
\label{ssec:classification}

After segmentation, the purpose of the classification task is to process the lesion patches and label them into non-dysplastic, dysplastic, and cancerous lesions. To distinguish between these three class labels, it is important to identify features such as texture and colour, exemplified in Fig. \ref{fig:data}. Classification is carried out using off-the-shelf VGG-16 \cite{simonyan_very_2014} and DenseNet \cite{huang_densely_2016} models. Those models were used with varying depth and width of layers in the classification head. Trained weights from ImageNet \cite{deng_imagenet_2009} were used to initialise the models and then fully trained on clinical images. We take advantage of the MIL technique by dividing our dataset into two broad groups/bags: cancerous (containing dysplastic (orange) and cancerous (red) lesions) and non-cancerous (containing non-dysplastic lesions), as shown in Fig. \ref{fig:data}.

Based on MIL, the network architecture learns from the backpropagation of two loss functions. First, $L_1$ compares predicted probabilities for each class N of every image in the bag B. As there are three classes, categorical cross-entropy is used as in (\ref{equ:finelabel}).

\begin{equation}\label{equ:finelabel}
	 L_1 = -\sum_{i=1}^{i=B}\sum_{j=1}^{j=N} {y_{ij}} \cdot {\log({\hat{y_{ij}}})}
\end{equation}

\begin{equation}\label{equ:baglabel}
	 L_2 = -\frac{1}{N}\sum_{i=1}^{i=N} {y_i} \cdot {\log{\hat{z_i}}} + (1-y_i)\cdot\log({1-\hat{z_i}})
\end{equation}

\begin{equation}\label{equ:bagprob}
	 \hat{z_i} = \frac{1}{B}\sum_{i=1}^{i=B} \sum_{j=1}^{j=N=2} \hat{y_{ij}}
\end{equation}

Second, binary cross entropy is defined in (\ref{equ:baglabel}). The predicted probability for a bag $\hat{z_i}$ is computed by combining the predictions $\hat{y_{ij}}$. For positive cases, i.e., the cancerous bag, average metrics are computed by adding predictions of dysplastic and cancerous, whereas for a negative bag, the average is computed between predictions of non-dysplastic items in the bag. (\ref{equ:bagprob}) shows how the probabilities for the positive class are computed, as the number of classes N = 2, i.e., the dysplastic and cancerous classes.

For training, the batch size is set to 32 patches of size $128$$\times$$128$ px and a bag size of 4. At first, VGG-16 \cite{simonyan_very_2014} and DenseNet \cite{huang_densely_2016} are fine-tuned on the clinical image patches and transfer learning is incorporated while training the MIL counterparts i.e., VGG-16 MIL (as shown in Fig. \ref{fig:block-diagram}B) and DenseNet-MIL. This is done by initiating trainable weights from clinical image training and further build-on with the loss functions. The novelty of the classification approach lies within (a) MIL infused training; (b) the newly created layer to merge probabilities for the bags; and (c) the transfer learning based approach that fine-tunes the off-the-shelf model on this dataset using its trainable weights to further improve results with MIL training.

\section{Results and Discussion}
\label{sec:results}

In this section, the results of the proposed detection, segmentation, and classification pipelines are reported. Table \ref{tab:segmentation-results} enumerates the evaluation results of the segmentation system. As described in Section \ref{ssec:data}, the enhanced model is evaluated with the default loss function along with the improved architecture Dice loss function, where the latter produced an average Dice score (\(DICE=\frac{2 |X \cap Y|}{|X| + |Y|})\) of 57.1\% and an overlap accuracy (i.e., the accuracy of bounding box detection at 25\% overlap) of 82.4\% on the unseen test dataset. 

It is worthy to mention that test subset comprises images of varying resolutions (e.g., small to high image size) and have been collected using different capture devices, lighting conditions, camera angles, hospital locations, and patient demographics, which provides a much needed heterogeneity in evaluating such models. In order to further illustrate the segmentation model capabilities, an online web application is developed using Gradio\footnote{https://huggingface.co/spaces/alsalemi/oed}. The reader can select a sample or upload an image to run the model and predict the lesion bounding box and mask. Further, the results of the model are benchmarked against related work in Table \ref{tab:segmentation-benchmark}. \textcolor{black}{In order to establish a fair comparison, the benchmarking was carried out using the same dataset and using the default parameters provided for each compared model.}

\begin{table}[!b]
    \centering
\caption{Segmentation evaluation metrics}
\label{tab:segmentation-results}
\resizebox{\columnwidth}{!}{\begin{tabular}{|l|c|c|c|} \hline 
            \textbf{Model}&\textbf{Mask Dice}&  \textbf{BBox F1}& \textbf{BBox Overlap Accuracy}\\ \hline 
            Mask R-CNN + Dice Loss&55.2\%&  60.5\%& 76.1\%\\ \hline 
            Transformer Mask R-CNN&52.6\%&  62.9\%& 76.4\%
\\ \hline
 \textbf{ Transformer Mask R-CNN + Dice Loss}& \textbf{57.1\%}& \textbf{69.2\%}&\textbf{82.4\%}\\\hline
\end{tabular}}
\end{table}

\begin{table}[!b]
    \centering
\caption{Benchmarking segmentation with related work}
\label{tab:segmentation-benchmark}
\resizebox{\columnwidth}{!}{\begin{tabular}{|l|l|c|} \hline 
            \textbf{Benchmarking Algorithm} &\textbf{Used In}& \textbf{Mask Dice}\\ \hline 
                       Mask R-CNN &\cite{anantharaman_utilizing_2018}& 52.1\%\\ \hline 
     U-Net&\cite{tanriver_automated_2021}& 28.5\%\\ \hline 
         
   \textbf{Transformer Mask R-CNN + Dice Loss}&\textbf{Proposed Pipeline}&\textbf{57.1\%}\\ \hline
\end{tabular}} 
\end{table}

\begin{figure}[!t]
\centering
\noindent\includegraphics[width=0.90\linewidth]{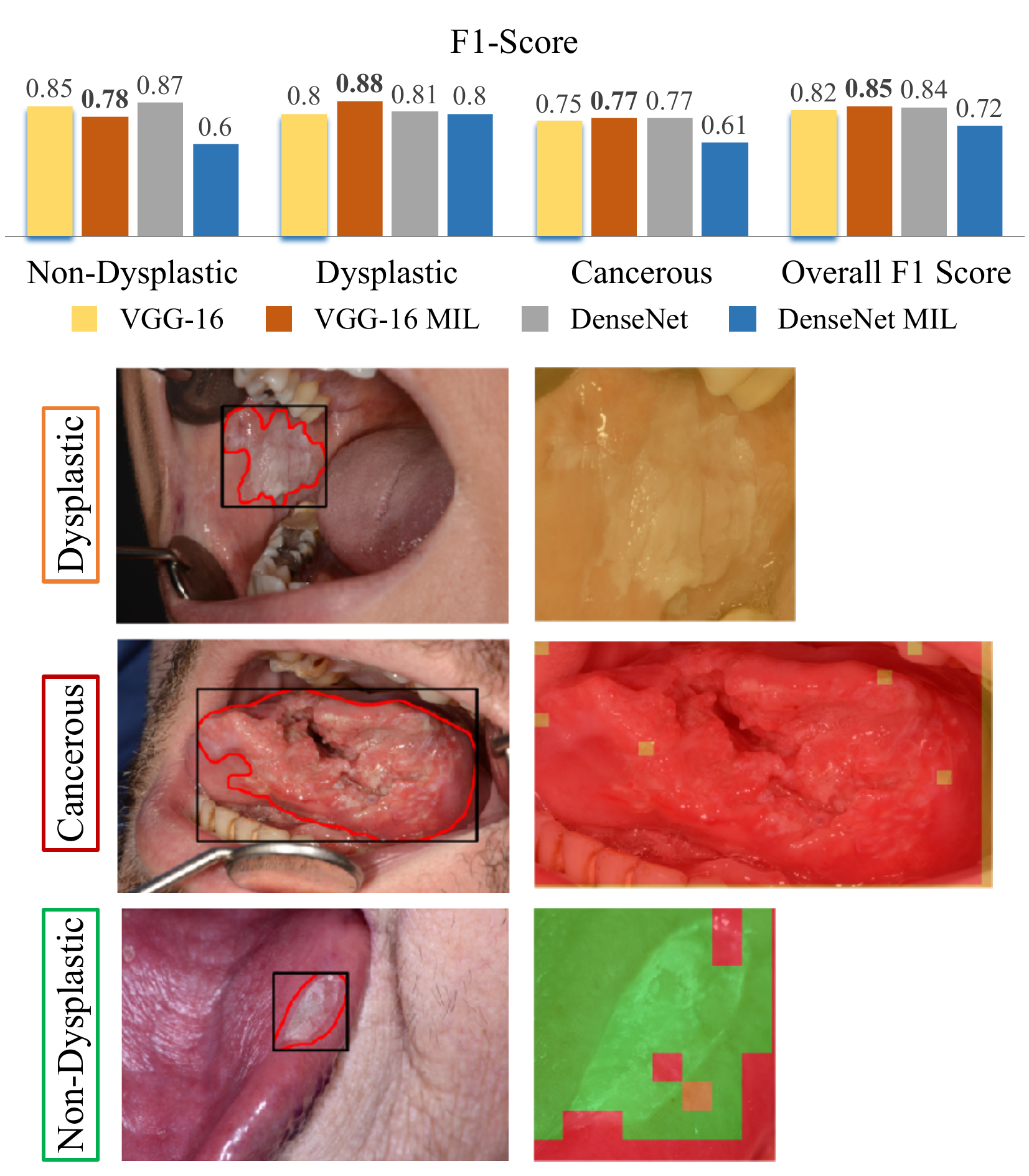}
\caption{Top: F1-score for VGG-16, VGG-16 MIL, DenseNet and DenseNet MIL, where MIL is the network architecture shown in Fig. \ref{fig:block-diagram}B. Bottom: Sample images classified by the VGG-16 MIL model.}
\label{fig:classification_results}
\end{figure}

Moreover, we evaluate our classification model, with a focus on the accurate grading of clinical images and mitigating false negatives. The F1-score is reported in the bar plot in Fig. \ref{fig:classification_results} for VGG-16, VGG-16 MIL (ours), DenseNet and DenseNet MIL (ours), where MIL is the network architecture shown in Fig. \ref{fig:block-diagram}B. F1-score \cite{achararit_artificial_2023} is plotted individually for all classes and an overall score is computed based on the test set as shown in Fig. \ref{fig:classification_results}. \textcolor{black}{Looking at both individial classes and overall performance, our MIL model with VGG-16 achieves the highest overall F1-score of 85.0\% and also for the class label dysplastic, which is 88.0\%.} 

\textcolor{black}{To compare the evaluated models' performances, false positive and false negative output is discussed. When visualising predicted class scores, as shown in Fig. \ref{fig:classification_results}, a greater number of mislabels can be seen in the non-dysplastic class, mainly on the boundaries and less on the lesion}. It is mainly due to fewer non-dysplastic images in the used dataset. Another reason for misclassification is the unlabelled background that is covered within the lesion while cropping the bounding box. The dysplastic lesion is predicted with 100\% accuracy and few mis-labels are encountered in the cancerous lesion. \textcolor{black}{For example, the VGG-16 MIL model achieved a sensitivity of 100\%, 99.0\%, and 87.0\% for dysplastic, cancerous, and non-dysplastic classes respectively, and a specificity of 98.0\%, 99.0\%, and 97.0\% for dysplastic, cancerous, and non-dysplastic classes respectively.}

The current pipeline harbours some limitations, e.g., the lack of a prognosis functionality, which will be addressed in future work. \textcolor{black}{Once the limitations are addressed and performance is improved, the work can be tested in a clinical setting to verify its efficacy.} The code used is publicly available\footnote{https://github.com/Precision-Vision/oed-classification-segmentation}. 

\section{Conclusions}
\label{sec:conclusions}

In this paper, a novel attention based OED detection, segmentation, and classification pipeline is proposed, taking advantage of a multi-source dataset. As a step towards early detection of head and neck cancer, the improved attention based Mask R-CNN segmentation architecture produces up to 82.0\% overlap accuracy score, while the MIL classification system achieves up to 85.0\% F1-score on the internal cohort validation set. Future work involves addressing the aforementioned limitations as an integrated pipeline, and employing endoscopic video data for automated early detection and prognosis.

\section{Compliance with Ethical Standards}
\label{sec:ethics}

This study was performed in line with the principles of the Declaration of Helsinki. Ethics approval was obtained from West of Scotland Research Ethics Service, REC reference 20/W/0017.

\textit{}\section{Author Contribution}
A. Alsalemi and A. Shakeel had equal contribution. S. E. A. Raza and S. A. Khurram had equal contribution.

\section{Acknowledgments}
\label{sec:acknowledgments}
This research is supported by Cancer Research UK Primer Award (EDDPMA-Nov21\textbackslash100017). Syed Ali Khurram is a shareholder in Histofy, an AI startup. The other authors declare no conflict of interest.

\bibliographystyle{IEEEtran}
\bibliography{references}

\begin{thebibliography}{10}
\providecommand{\url}[1]{#1}
\csname url@samestyle\endcsname
\providecommand{\newblock}{\relax}
\providecommand{\bibinfo}[2]{#2}
\providecommand{\BIBentrySTDinterwordspacing}{\spaceskip=0pt\relax}
\providecommand{\BIBentryALTinterwordstretchfactor}{4}
\providecommand{\BIBentryALTinterwordspacing}{\spaceskip=\fontdimen2\font plus
\BIBentryALTinterwordstretchfactor\fontdimen3\font minus \fontdimen4\font\relax}
\providecommand{\BIBforeignlanguage}[2]{{%
\expandafter\ifx\csname l@#1\endcsname\relax
\typeout{** WARNING: IEEEtran.bst: No hyphenation pattern has been}%
\typeout{** loaded for the language `#1'. Using the pattern for}%
\typeout{** the default language instead.}%
\else
\language=\csname l@#1\endcsname
\fi
#2}}
\providecommand{\BIBdecl}{\relax}
\BIBdecl

\bibitem{nevanpaa_malignant_2022}
T.~T. Nevanpää \emph{et~al.}, ``Malignant transformation of oral epithelial dysplasia in {Southwest} {Finland},'' \emph{Scientific Reports}, May 2022.

\bibitem{noauthor_head_2019}
\BIBentryALTinterwordspacing
``Head and neck cancers statistics,'' 2019. [Online]. Available: \url{https://www.cancerresearchuk.org/health-professional/cancer-statistics/statistics-by-cancer-type/head-and-neck-cancers}
\BIBentrySTDinterwordspacing

\bibitem{gormley_reviewing_2022}
M.~Gormley \emph{et~al.}, ``Reviewing the epidemiology of head and neck cancer: definitions, trends and risk factors,'' \emph{British Dental Journal}, Nov. 2022.

\bibitem{mahmood_prediction_2022}
H.~Mahmood,  \emph{et~al.}, ``Prediction of malignant transformation and recurrence of oral epithelial dysplasia using architectural and cytological feature specific prognostic models,'' \emph{Modern Pathology}, Sep. 2022.

\bibitem{camalan_convolutional_2021}
S.~Camalan \emph{et~al.}, ``Convolutional {Neural} {Network}-{Based} {Clinical} {Predictors} of {Oral} {Dysplasia}: {Class} {Activation} {Map} {Analysis} of {Deep} {Learning} {Results},'' \emph{Cancers}, Mar. 2021.

\bibitem{shephard_fully_2023}
A.~J. Shephard \emph{et~al.}, ``A {Fully} {Automated} and {Explainable} {Algorithm} for the {Prediction} of {Malignant} {Transformation} in {Oral} {Epithelial} {Dysplasia},'' \emph{arXiv preprint arXiv:2307.03757}, 2023.

\bibitem{mahmood_artificial_2021}
H.~Mahmood \emph{et~al.}, ``Artificial {Intelligence}-based methods in head and neck cancer diagnosis: an overview,'' \emph{British Journal of Cancer}, Jun. 2021.

\bibitem{anantharaman_utilizing_2018}
R.~Anantharaman \emph{et~al.}, ``Utilizing {Mask} {R}-{CNN} for {Detection} and {Segmentation} of {Oral} {Diseases},'' in \emph{2018 {IEEE} {BIBM}}, Dec. 2018.

\bibitem{fu_deep_2020}
Q.~Fu \emph{et~al.}, ``A deep learning algorithm for detection of oral cavity squamous cell carcinoma from photographic images: {A} retrospective study,'' \emph{EClinicalMedicine}, Oct. 2020.

\bibitem{he_mask_2018}
K.~He \emph{et~al.}, ``Mask {R}-{CNN},'' Jan. 2018, arXiv:1703.06870 [cs].

\bibitem{jadon_survey_2020}
S.~Jadon, ``A survey of loss functions for semantic segmentation,'' in \emph{2020 {IEEE} {CIBCB}}, Oct. 2020.

\bibitem{li_benchmarking_2021}
Y.~Li \emph{et~al.}, ``Benchmarking {Detection} {Transfer} {Learning} with {Vision} {Transformers},'' Nov. 2021, arXiv:2111.11429 [cs].

\bibitem{simonyan_very_2014}
K.~Simonyan and A.~Zisserman, ``Very {Deep} {Convolutional} {Networks} for {Large}-{Scale} {Image} {Recognition},'' \emph{arXiv:1409.1556}, 2014.

\bibitem{huang_densely_2016}
G.~Huang \emph{et~al.}, ``Densely {Connected} {Convolutional} {Networks},'' \emph{arXiv:1608.06993}, 2016.

\bibitem{deng_imagenet_2009}
J.~Deng \emph{et~al.}, ``{ImageNet}: {A} large-scale hierarchical image database,'' in \emph{2009 {IEEE} {CVPR}}, Jun. 2009.

\bibitem{tanriver_automated_2021}
G.~Tanriver \emph{et~al.}, ``Automated {Detection} and {Classification} of {Oral} {Lesions} {Using} {Deep} {Learning} to {Detect} {Oral} {Potentially} {Malignant} {Disorders},'' \emph{Cancers}, Jan. 2021.

\bibitem{achararit_artificial_2023}
P.~Achararit \emph{et~al.}, ``Artificial {Intelligence}-{Based} {Diagnosis} of {Oral} {Lichen} {Planus} {Using} {Deep} {Convolutional} {Neural} {Networks},'' \emph{European Journal of Dentistry}, Jan. 2023.

\end{thebibliography}

\end{document}